\RequirePackage{fix-cm}
\documentclass[smallextended]{svjour3}       % onecolumn (second format)
\smartqed  % flush right qed marks, e.g. at end of proof
\usepackage{graphicx}
\usepackage{subfigure}
\usepackage{amsmath}
\usepackage{booktabs}
\usepackage{longtable}
\usepackage{setspace}
\usepackage{epstopdf}
\usepackage{multirow}
\usepackage{changepage}
\usepackage{threeparttable}
\usepackage[misc]{ifsym}
\usepackage{amsfonts,amssymb}
\usepackage{amsfonts}
\usepackage{multirow}
\begin{document}

\title{HEU Emotion: A Large-scale Database for Multi-modal Emotion Recognition in the Wild%\thanks{Grants or other notes
}


\author{
        Jing Chen \textsuperscript{1} \and Chenhui Wang \textsuperscript{2} \and Kejun Wang \textsuperscript{1} \and
        Chaoqun Yin \textsuperscript{1} \and Cong Zhao \textsuperscript{1} \and Tao Xu \textsuperscript{1} \and Xinyi Zhang \textsuperscript{1} \and Ziqiang Huang
\textsuperscript{1}        \and Meichen Liu \textsuperscript{1} \and Tao Yang \textsuperscript{1}%etc.
}

\institute{
\Letter Kejun Wang\\
\email{heukejun@126.com}\\
%Tel.: +86-21-34206291\\
\Letter Chenhui Wang\\
\email{harbin0451@g.ucla.edu}\\
Jing Chen\\
\email{jing\_chen@hrbeu.edu.cn}\\
\at
 {1} College of Intelligent Systems Science and Engineering, Harbin Engineering University, Harbin, China.
 \at
 {2} UCLA Department of Statistics, Los Angeles, CA.\\
}

%\institute{ Kejun Wang  %\\
             % \email{heukejun@126.com}            \\
              %\Letter Chenhui Wang  \\
             % \email{harbin0451@g.ucla.edu}            \\

%\at
% {1} College of Automation, Harbin Engineering University, Harbin, China.
% \at
% {2} UCLA Department of Statistics, Los Angeles, CA.\\
%}

\date{Received: date / Accepted: date}

\maketitle

\begin{abstract}
The study of affective computing in the wild setting is underpinned by databases. Existing multimodal emotion databases in the real-world conditions are few and small, with a limited number of subjects and expressed in a single language. To meet this requirement, we collected, annotated, and prepared to release a new natural state video database (called HEU Emotion). HEU Emotion contains a total of 19,004 video clips, which is divided into two parts according to the data source. The first part contains videos downloaded from Tumblr, Google, and Giphy, including 10 emotions and two modalities (facial expression and body posture). The second part includes corpus taken manually from movies, TV series, and variety shows, consisting of 10 emotions and three modalities (facial expression, body posture, and emotional speech). HEU Emotion is by far the most extensive multi-modal emotional database with 9,951 subjects. In order to provide a benchmark for emotion recognition, we used many conventional machine learning and deep learning methods to evaluate HEU Emotion. We proposed a Multi-modal Attention module to fuse multi-modal features adaptively. After multi-modal fusion, the recognition accuracies for the two parts increased by 2.19\% and 4.01\% respectively over those of single-modal facial expression recognition.
\end{abstract}

\section{Introduction}
\label{intro}
Emotion recognition is intended to perceive a person's emotional state through various information channels such as face, speech, physiological signals, and the like. Both rhythm and substance are flourishing in parallel when a person expresses his or her emotional needs. It means that people arouse others' senses in a variety of ways. Current automatic emotion recognition systems, however, are still stuck in identifying a single feature. They lack all emotions and social skills that building intelligent human-computer interaction needs. There are two main reasons: on one hand, the operating environment is often uncontrollable, such as complex background, illumination change, change of camera angle, age, race, and gender of the user. On the other hand, complexity and diversity of emotions make it challenging to realize intelligent human-computer interaction. Constructing an emotional database under the above uncontrollable environment and including as many expressive categories as possible is an essential step in promoting the practical application of emotion recognition.

Many popular emotion recognition databases are collected in controlled laboratory environments. The subjects are called upon to make certain expressions in lab scenarios. These deliberately generated expressions are not tangible emotions. Moreover, there are some limitations, including a plain background, a single light, and few subjects. Therefore, the actual application effect of an emotion recognition system developed based on these databases is not satisfactory. With the rise of various social platforms and networks, millions of users upload and share their photos and video clips every day. Shooting scenes of these videos are the real application environment of an automatic emotion recognition system. Although some video clips are clipped from movies, they are closer to the real world than the previous databases from lab conditions.

In recent years, as more and more emotional databases \cite{8013713,8099760,dhall2011static,goodfellow2013challenges} in the wild have been collected and annotated, research on automatic algorithms for emotion recognition in the real world has gradually increased. These databases contain different backgrounds and a large number of testees. However, most of the databases are facial expressions and single static images. Consequently, it is difficult for researchers to utilize other emotional expression channels and to learn emotional changes. The process of emotional change is often continuous and time-correlated. Therefore, the information of dynamic expression sequence is more comprehensive and specific than that of a single static image in capturing emotions. Psychology professor Mehrabian pointed out that expression, speech, and language accounted for 55\%, 38\%, and 7\% in emotional interaction, respectively \cite{mehrabian2008communication}. The expressions are divided into facial expressions and body postures. One or more movements of facial muscle produce facial expressions. At present, good classification accuracies of basic emotions in the natural environment have been achieved by using face images (more than 70\% \cite{fan2018multi,georgescu2019local} on the RAF-DB \cite{8099760} dataset). Body postures are considered to be another nonverbal communication method of human emotions. They are often understood as changes in the movement of the head, limbs, and other parts of the body. Relative to collecting face, low-resolution collection facilities are enough to acquire postures. As a perfect complement of emotion recognition, the analysis of body movement data has recently become more common. Speech is another critical channel for emotion recognition. Human beings can capture the state of the other's emotions by listening to the speech and perceiving the speaker's modal particle and phonetic tone. According to reports, some emotions (such as sadness and fear) are more easily distinguished from audio signals than from visual appearance \cite{de1997facial}. Emotions are expressed through multiple channels. The emotions collected from the real world are difficult to classify through one channel. In the case of single-modality studies, information is often insufficient, and the classification results are susceptible to various external factors, such as face or body occlusion, and noise. The McGurk effect \cite{mcgurk1976hearing} reveals that different organs are automatically unintentionally combined to process the information when brain is sensing. The absent or inaccuracy information can lead to deviations of brain's understanding of external information. That is why multimodal technologies have become more prevalent in automatic emotion recognition recently.

As mentioned in Sect.\ref{Related work}, over the past decades, many organizations have made great efforts to establish multimodal emotion databases. However, there are still three main problems: (1) The number of samples of the multimodal emotion database in the natural state is small. There is a big gap compared to the enormous demand for practical applications. (2) The number of subjects in these datasets is still small, as shown in Table \ref{table1}, with a maximum of 527 subjects. The limited number of subjects hinders the study of identity-independent emotion analysis and recognition. (3) The languages of the emotional databases in the natural state in Table \ref{table1} are single, and the cultural backgrounds of the subjects in one dataset are the same. Cultural divergences may lead to differences in emotional expression. The actual application effects of systems developed based on these databases can vary greatly depending on the attributes of the user.

In order to overcome the above problems in the multimodal emotion databases, we collected and annotated a sizeable multi-modal emotion database (HEU Emotion) in the wild environment. The advantages of HEU Emotion are as follows:
\begin{itemize}
\item First of all, it is by far the largest multimodal emotion database collected in the natural state. It includes 16,569 video clips downloaded from different websites (Tumblr, Google, Giphy), as well as 2,435 corpora selected from movies, TV series, and live videos.

\item Secondly, there are 9,951 subjects in HEU Emotion. A massive increase in the number of subjects can significantly reduce the impact of identity information on emotion analysis and recognition. Compared to the existing multimodal emotional datasets shown in Table \ref{table1}, HEU Emotion has the largest number of subjects.

\item Finally, there are many speakers from different cultural backgrounds, such as Chinese, Americans, Thais, Koreans, etc. In most cases, they speak their native language. Therefore HEU Emotion is an emotional database with multiple languages. Besides, in order to enrich the emotional categories, we annotated three emotions (disappointed, confused, and bored) in addition to the basic seven emotions.
\end{itemize}

\section{Related work}
\label{Related work}
\begin{table*}[htbp] %开始一个表格environment，表格的位置是h,here。
\newcommand{\tabincell}[2]{\begin{tabular}{@{}#1@{}}#2\end{tabular}}
\centering
\caption{Overview of existing multimodal emotional databases:  Subs denotes the number of subjects collected in the databases; Cond represents the collection environment; Samples shows the number of corpus; Language indicates the language used in the databases; Data demonstrates the types of multimodal data recorded; Dist indicates the number of categorical emotion states.}\label{table1} % 显示表格的标题
\begin{tabular}{ccccccc} %设置了每一列的宽度，强制转换。

\hline
%\toprule[1pt]
%Format & Extension & Description \\ % 用&来分隔单元格的内容 \\ 表示进入下一行
%\hline %画一个横线，下面的就都是一样了，这里一共有4行内容
%
Databases& Subs& Cond& Samples& Language& Data& Dist\\
\hline
FABO \cite{gunes2006bimodal}& 23& Lab& 246& -& Face and posture& 9 \\
%\hline
RAVDESS \cite{livingstone2018ryerson}& 24& Lab& 7,356& English& Face and speech& 7\\
%\hline
RAMAS \cite{perepelkina2018ramas}& 10& Lab& 581& Russian& Face, speech, posture and ECG& 6\\
%\hline
Sapinski et al. \cite{sapinski2018multimodal}& 16& Lab& 560& Polish& Face, speech and posture& 7\\
%\hline
NNIME \cite{chou2017nnime}& 44& Lab& -& Chinese& Face, speech and ECG& 6\\
%\hline
SEMAINE \cite{mckeown2011semaine}& 150& Lab& 959& English& Face and speech& 27\\
%\hline
SAVEE \cite{jackson2014surrey}& 4& Lab& 480& English& Face and speech& 7 \\
%\hline
eNTERFACE'05 \cite{martin2006enterface}& 42& Lab& -& English& Face and speech& 7\\
%\hline
RML \cite{jackson2014}& -& Lab& 720& multilingual& Face and speech& 6 \\
%\hline
IEMOCAP \cite{busso2008iemocap}& 10& Lab& 3,060& English& Face, speech and gesture& 10\\
\hline
Yu et al. \cite{yu2001emotion}& -& Wild& 721& Chinese& Face and speech& 4\\
%\hline
SAFE \cite{clavel2006safe}& 400& Wild& 4,073& multilingual& Face and speech& 4\\
%\hline
Fiction \cite{vasilescu2004fiction}& 28& Wild& -& English& Face and speech& 4\\
%\hline
MELD \cite{poria2018meld}& -& Wild& 1,433& English& Face, speech and textual& 7\\
%\hline
CHEAVD \cite{li2017cheavd}& 238& Wild& 2,322& Chinese& Face and speech& 6\\
%\hline
CHEAVD 2.0 \cite{li2018mec}& 527& Wild& 7,030& Chinese& Face and speech& 8\\
%\hline
AFEW \cite{6200254}& 330& Wild& 1,809& English& Face, speech and posture& 7\\
\hline
HEU-part1& 8,984& Wild& 16,569&-& Face and posture& 10\\
%\hline
HEU-part2& 967& Wild& 2,435& multilingual& Face, speech and posture& 10\\
\hline
%\bottomrule[1pt]
\end{tabular}
\end{table*}
In order to extend the research to the real environment, many research institutions have created emotion recognition datasets in the real environment. FER2013 \cite{goodfellow2013challenges} is a database of the ICML2013 facial expression recognition Challenge. RAF-DB \cite{8099760} dataset is a real-world facial expression database, which is widely used at present. Nevertheless, these datasets are static images that ignore dynamic emotional changes and focus only on facial expressions. They do not make use of the characteristics of multiple ways of emotional expression. This section focuses on datasets that contain dynamic emotional changes and contain multiple emotional modes.

\textbf{FABO} \cite{gunes2006bimodal} database used two cameras to capture simultaneously facial expressions and gestures. The shooting angle was positive, and the background was a blue static setting. At the same time, in order to reduce the impact of illumination changes, the researchers built an artificial light source environment. The subjects were not directly exposed to the light source. At the time of the shooting, the researcher asked the subjects to make specific emotions and perform a combination of different facial expressions and upper body gestures. This bimodal database consists of 23 subjects, 11 males and 12 females, who are between 18 and 50 years old. They come from different countries and regions, such as Europe, the Middle East, Latin America, Asia, and Australia. The researchers filmed nine emotions: angry, scared, surprised, happy, disgusted, bored, worried, sad, and uncertain.

\textbf{RAVDESS} \cite{livingstone2018ryerson} includes 60 speeches and 44 songs with emotions (including fear, sadness, surprise, happy, anger, disgust, and neutral), which were recorded by 24 professional actors. There were three forms of work recorded by each actor: audio-visual (AV), video-only (VO), and audio-only (AO). Recordings were recorded in professional studios, with only actors and green screens visible in the lens. To ensure that the camera was able to capture the actor's head and shoulder, the height of the camera was adjusted at any time. Full-spectrum illumination was provided in the studio, illuminated by ceiling fluorescent light and three 28W 5200k CRI 82 bulbs. These settings could minimize facial shadows.

\textbf{RAMAS} \cite{perepelkina2018ramas}  is the first Russian multimodal emotional database collected by Neurodata Lab LLC. Ten semi-professional actors (5 males and 5 females, aged 18-28, native Russian) were involved in the data collection. The collectors thought that professional drama actors might use stereotypes of action patterns. Accordingly, semi-professional actors were more suitable for performing actions in emotional situations. The semi-professional actor expressed one of the basic emotions (anger, sadness, disgust, happiness, fear, and surprise) in the set scene. Various data such as audio, motion capture, close-up and panoramic video, and physiological data were gathered during recording.

\textbf{Sapinski et al.} \cite{sapinski2018multimodal} published an emotional database in the Polish language, consisting of three forms: facial expressions, body movements and gestures, and speech. The recordings were recorded in the rehearsal room of $Teatr\ Nowy\ im$ by 16 professional actors (8 males and 8 females aged 25 to 64).
Recordings were performed in a quiet, well-lit environment with a green background. In order to keep the actor's face in the picture and compensate for any movement in the emotional expression process, a medium shot was used. In the case of Kinect recording, the entire body was in the frame, including the legs.

\textbf{NNIME} \cite{chou2017nnime} is a recording of dyadic spoken interactions. Forty-four (24 females and 20 males) subjects from the National Taiwan University of the Arts Drama (NTUA) took part in the recording. Every two people were split into one group (7 female-female, 10 female-male, and 5 male-male pairs). Each pair was instructed to perform spontaneously a short scene of about 3 minutes. The overall performance was to provide evidence for one of six pre-specified emotions (anger, sadness, happiness, frustration, neutrality, and surprise).

\textbf{SEMAINE} \cite{mckeown2011semaine} is a multimodal database of emotional conversations between human subjects and computer conversational agents, which are collected by McKeown et al. High-quality recording was made by 5 high frame rate, high-resolution cameras, and 4 microphones. A total of 959 conversations with 150 participants and 4 Sensitive Artificial Listener (SAL) characters were recorded, each lasting approximately 5 minutes. Each clip was traced to 27 related categories by 6-8 raters. Furthermore, there were four main types of basic emotions, epistemic states, interaction process analysis, and validity.

\textbf{SAVEE} \cite{jackson2014surrey} is a recording of four male subjects between the ages of 27 and 31, containing a total of 480 samples of 7 basic emotions. It used an inductive approach to collect data. Expressions and recordings were recorded when subjects watched video clips and text on the monitor.

\textbf{eNTERFACE'05} \cite{martin2006enterface} has audio-visual clips recorded in a laboratory environment. Forty-two subjects from different nationalities participated in the recording, and the recorded language was English. The dark gray background was used for the acquisition, and the captured images contained only heads.

\textbf{RML} \cite{jackson2014} was collected by Ryerson Laboratories and contained 720 samples of six basic expressions. Eight subjects who participated in the recordings spoke various languages. The recording was done in a no-noise atmosphere with a simple gray-green background and a digital camera to capture the videos.

\textbf{IEMOCAP} \cite{busso2008iemocap} is an action, multi-modal, and multi-peak database collected by the University of Southern California's Sail Lab. It contains approximately 12 hours of audiovisual data, including video, voice, facial motion capture, and text transcription. Participants performed impromptu performances or scripting scenarios. IEMOCAP was annotated by many annotators into category tags such as anger, happiness, sadness, neutrality, and dimension labels such as valence, activation, and dominance.

\textbf{Yu et al.} \cite{yu2001emotion} collected 721 phrases from Chinese movies and TV series, including 4 emotions (anger, happiness, sadness and neutral).

\textbf{SAFE} \cite{clavel2006safe} and \textbf{Fiction} \cite{vasilescu2004fiction} focuses on extreme emotions in abnormal situations. Unlike other databases, the emotions of these two databases fell into four categories: fear, negative emotions, neutral emotions, and positive emotions.

\textbf{MELD} \cite{poria2018meld} was evolved from the EmotionLines dataset \cite{chen2018emotionlines}. EmotionLines only contains conversations from the TV series ``Friends". MELD is a multimodal emotional conversation dataset, containing audio, visual, and text modalities. Since data were obtained from only one TV series, the number of participants was limited, and 84\% of the sessions were obtained by 6 starring actors.

\textbf{CHEAVD} \cite{li2017cheavd} contains 140 minutes of emotional episodes from movies, TV shows, and talk shows. There are a total of 26 atypical emotional states, and a variety of emotional tags and fake/suppressed emotional tags. However, the emotional categories used in the final experiment were only six basic emotions. Moreover, the number is exceptionally uneven, $neutral : surprised: disgust = 19.6: 1.6: 1$.

\textbf{CHEAVD 2.0} \cite{li2018mec} is an extension of CHEAVD. In addition to the six basic emotions, it adds worried and anxious. As in \cite{li2017cheavd}, the database also has an extremely unbalanced number of sample, $neutral: surprised: disgust = 10 : 1.2: 1$. Also, from the confusion matrix of the given baseline, it could be seen that other categories except for anger, happiness, and neutral were challenging to identify. Especially, the recognition accuracies of surprise and disgust were zero.

\textbf{AFEW} \cite{6200254} is an emotional database in the wild environment proposed by A. Dhall et al. Since 2013, it has been used as a database for the Emotion Recognition in the wild Challenge (EmotiW). It was captured from 54 Hollywood movies, including 1,809 video clips with various head poses,
occlusions, and different lighting. AFEW is a multimodal database of seven basic emotion categories. Researchers are encouraged to use a variety of modalities, such as facial expressions, postures, and audio signals. Blocked by the organizer, only 773 training set samples and 383 verification set samples are publicly available.

Table \ref{table1} summarizes the characteristics of the reviewed databases in categorical model.

From the above review, it can be seen that many audio-visual databases were collected in a laboratory environment. The non-laboratory databases, SAFE and Fiction, focus on extreme expressions and are not conventional emotion recognition. CHEAVD and CHEAVD 2.0 are collected from movies and TV series, but their samples are only in the Chinese language. AFEW only provides English samples. Besides, these datasets have the most natural expressions and laughter expressions, and the amount of other expressions which are not easily distinguishable is even less. Due to the current small scale of emotional datasets in the natural environment, the single language, and the uneven distribution of various emotional samples, we established a large-scale, multiple languages and relatively balanced database HEU Emotion. Additionally, our HEU Emotion was recorded in the natural environment. Although it contains many noises, it is more in line with the real-world application environment and can help to further improve the generalization and robustness of the emotion recognition system.

\section{Heu emotion database}
\label{sec:Heu emotion database}
\begin{figure*}[htbp]
\centering
\includegraphics[width=15cm]{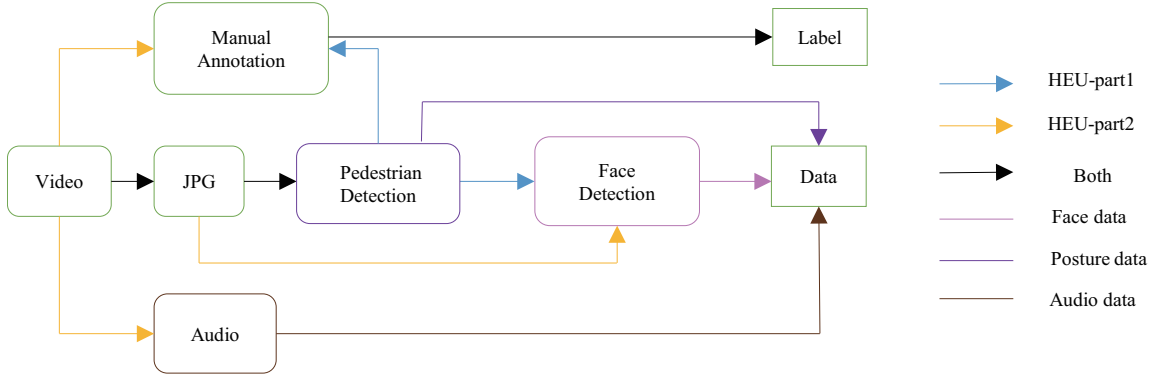}
\caption{The process of building the HEU Emotion database}
\label{fig1}
\end{figure*}

HEU Emotion is currently the largest video-based multimodal emotion recognition database in the wild. In order to make the benchmark test closer to the actual application, we downloaded the relevant video clips by retrieving the emotion-related keywords from search engines such as Tumblr, Google, and Giphy. Meanwhile, the video clips were manually chosen from online videos such as movies, TV shows, and variety shows. Data of HEU Emotion part1 (HEU-part1) contained numerous non-personal information like drawing, graphics, or non-human objects which needed to be filtered out by an automatic algorithm. All clips extracted manually in HEU Emotion Part2 (HEU-part2) contained characters and therefore could be labeled directly. Figure \ref{fig1} shows the process of database construction, and the information represented by each color is given on the right side of the graph. Details will be presented in the following sections.

\subsection{Collecting Data}\label{subsec:collecting}
Firstly we created a query list of emotional keywords (e.g., nervous, disgusted, happy, sad, fear, anger, scared, surprised, bored, confused, disappointed, frustrated, etc). Three search engines (Tumblr, Google, Giphy) were queried with these emotion related tags, and URLs of videos were stored in a document. Then, 47,450 original video clips were obtained by batch download using the automatic downloader. Since a high percentage of the results returned by our query words already contained neutral expression videos, no separate query was made to obtain additional neutral expressions. The 3500+ original clips in HEU-part2 were picked out by 5 staff members. Manually edited videos follow the following rules: (1) in each video, the performer has a single expression. That is, only one expression appears, and the other expressions do not appear as far as possible. (2) the shots in the video should be kept on the actors who express their emotions as far as possible, to avoid switching back and forth. (3) A long video can be divided into several parts. However, each part presents a different stage of emotional expression. (4) multiple performers can be included in the same frame, but in most instances all try to express the same emotion.

\subsection{Extracting Frames}
\label{subsec:extracting frames}
After going to gain the video clips, we did the relevant processing work on the videos. The FFmpeg of OpenCV was used to get the JPG image from each video. The frame rate was given according to that of each video instead of setting a fixed one. Generally, humans can complete the entire process of emotional change and reach a peak within 10 seconds. Consequently, to reduce redundant information and computational burden, we re-edited the videos with lengths longer than 10 seconds.

\subsection{Automatic Filtering by Pedestrian Detection}
\label{subsec:automatic filtering by pedestrain detection}
\begin{figure}[htbp]
\centering
\includegraphics[width=8.5cm]{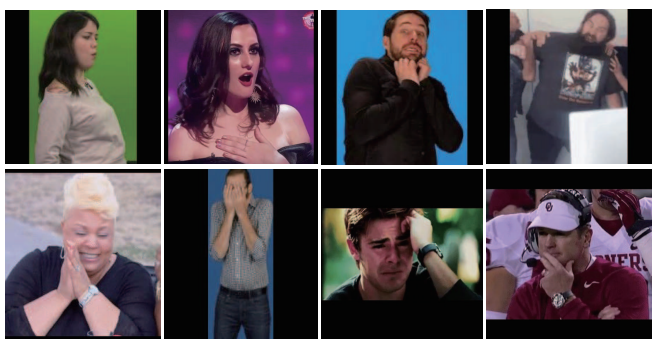}
\caption{The normalization samples of body posture}
\label{fig2}
\end{figure}

Considering that the video data which were obtained from Internet contained non-human objects, the video frames were filtered by YOLO V3 \cite{redmon2018yolov3}. Among them, a frame was defined as a solid frame in which people can be detected. A clip containing of valid frames was defined as a sound clip, and invalid clips were directly deleted. We picked out characters from the original pictures under the bbox positioned by YOLO V3. When there was more than one character in the video, the emotion was often expressed by the one who took up a larger area. We calculated the area of the box where each character was located according to the position of the given bbox. Then, to make sure that only one character was in the image, the one with the largest area was presented after pedestrian detection. Since the sizes of images obtained directly were different, we performed a normalization process. In consequence of the irregular sizes of raw images, we could not find a suitable size to fit all the images, so we put the images into squares. Moreover, the background of the original image was generally complicated, and thus, we did not expand the cutting position on the original image. Instead, the difference between the length and width was calculated. After that, one-half of difference was expanded on both sides of the small side to obtain a rectangular picture. Results of normalization are shown in Fig. \ref{fig2}.

\subsection{Obtaining Facial Expression Data}\label{subsec:obtaining facial expression data}
\begin{figure}[htbp]
\centering
\includegraphics[width=8.5cm]{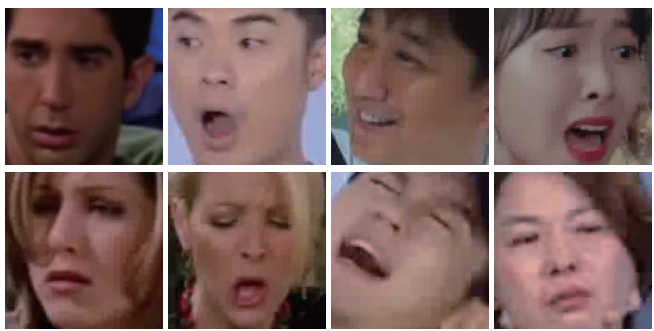}
\caption{Samples of Face Images Detected by Libface-detection}
\label{fig3}
\end{figure}

\begin{figure}[htbp]
\centering
\includegraphics[width=8.5cm]{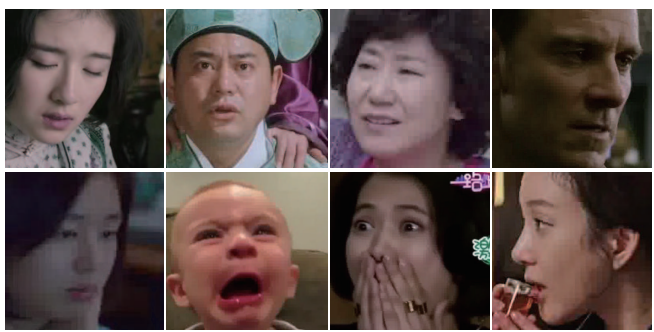}
\caption{ Samples of Face Images Detected by MTCNN}
\label{fig4}
\end{figure}

In the face detection section, we used two efficient face detection methods, MTCNN \cite{zhang2016joint} and libface-detection (libface)\footnote{https://github.com/ShiqiYu/libfacedetection} to achieve the effect of mutual complementation and to reduce missed detection. Figure \ref{fig3} shows the images detected by libface, which only contained the faces. The length-width ratios of the images detected by MTCNN were not uniform. In order to facilitate the training of the model, the detected images were processed as follow. The values of the two sides were respectively calculated in the light of the given position of the box. The starting position of the smaller side was unchanged, and the ending position was calculated on the basis of the length of the longer side and the starting position.

\begin{equation}
h = bbox[0][1] - bbox[0][0]
\end{equation}
\begin{equation}
w = bbox[1][0] - bbox[0][0]
\end{equation}
\begin{equation}
\begin{aligned}
&bbox_{new}[1][0] = bbox[0][0] + h, ~{if~ h > w} \\
&bbox_{new}[1][1] = bbox[0][1] + h, ~{if~ h > w}
\end{aligned}
\end{equation}
\begin{equation}
\begin{aligned}
&bbox_{new}[0][1] = bbox[0][0] + w, ~{if ~h < w} \\
&bbox_{new}[1][1] = bbox[1][0] + w, ~{if ~h < w}
\end{aligned}
\end{equation}

where bbox is the coordinates of the detected face. Bbox[0][0] is the initial position of the $x$ and $y$ axes of the picture; Bbox[0][1] is the initial position of the $x$-axis and the end position of the $y$-axis; Bbox[1][0] is the end position of the $x$-axis and the initial position of the $y$-axis; Bbox [1] [1] is the end position of $x$-axis and $y$-axis. Therefore, $h$ is the length in the $y$-axis direction and $w$ is the length in the $x$-axis direction. Images detected by MTCNN included not only face but also head. Samples are presented in Fig.\ref{fig4}. Besides, the data of the HEU-part1 were more complicated, and face detections were based on the images detected by \cite{redmon2018yolov3}. The data in the HEU-part2 were ideal, and they were directly detected on the JPG images. In practical application, libface has better recognition ability for a variety of facial angles and occlusion. But many images are misdetected as local organs (such as nose, ears, hands, etc.). The final face image is mainly the face image detected by MTCNN. When MTCNN can not detect any face, the data detected by libface is used, and the bad image is deleted manually.

\subsection{Database Annotation}\label{subsec:database annotation}
 The keywords used when downloading HEU-part1 depend on the information annotated by the user at the time of upload, which is not exactly consistent with the real emotional state. That means that all data needs to be re-tagged. Annotating nearly 45,000 videos is a challenging and time-consuming task. In the process of manual tailoring of HEU-part2, collectors were asked to cut videos according to the emotional state. However, the deviation of human emotional judgment and the different sensitivity of each person to emotion exists. In order to avoid the error caused by personal judgment, the annotation work was carried out jointly by multiple annotators. The annotation process is wholly blind and independent. There were a total of 15 tagging staff, and they were taught emotional psychology for a week before the tagging work began, including the definition of each emotional category, main features and some examples. Fifteen annotators tagged all the video clips independently and divided them into the most obvious of the 10 categories. We counted the annotated data, marked it by voting, and selected the one with the highest number of votes as its label. If the number of votes in several categories was very similar. One of them did not appear to be significantly higher than the other categories. We put the clip as a mixed emotion in another class. Because of limited time and workforce, there was not any further processing of the compound expression.

\section{Statistics}
\label{sec:statistics}
\begin{figure}[htbp]
\centering
\subfigure []{
\includegraphics[width=4cm]{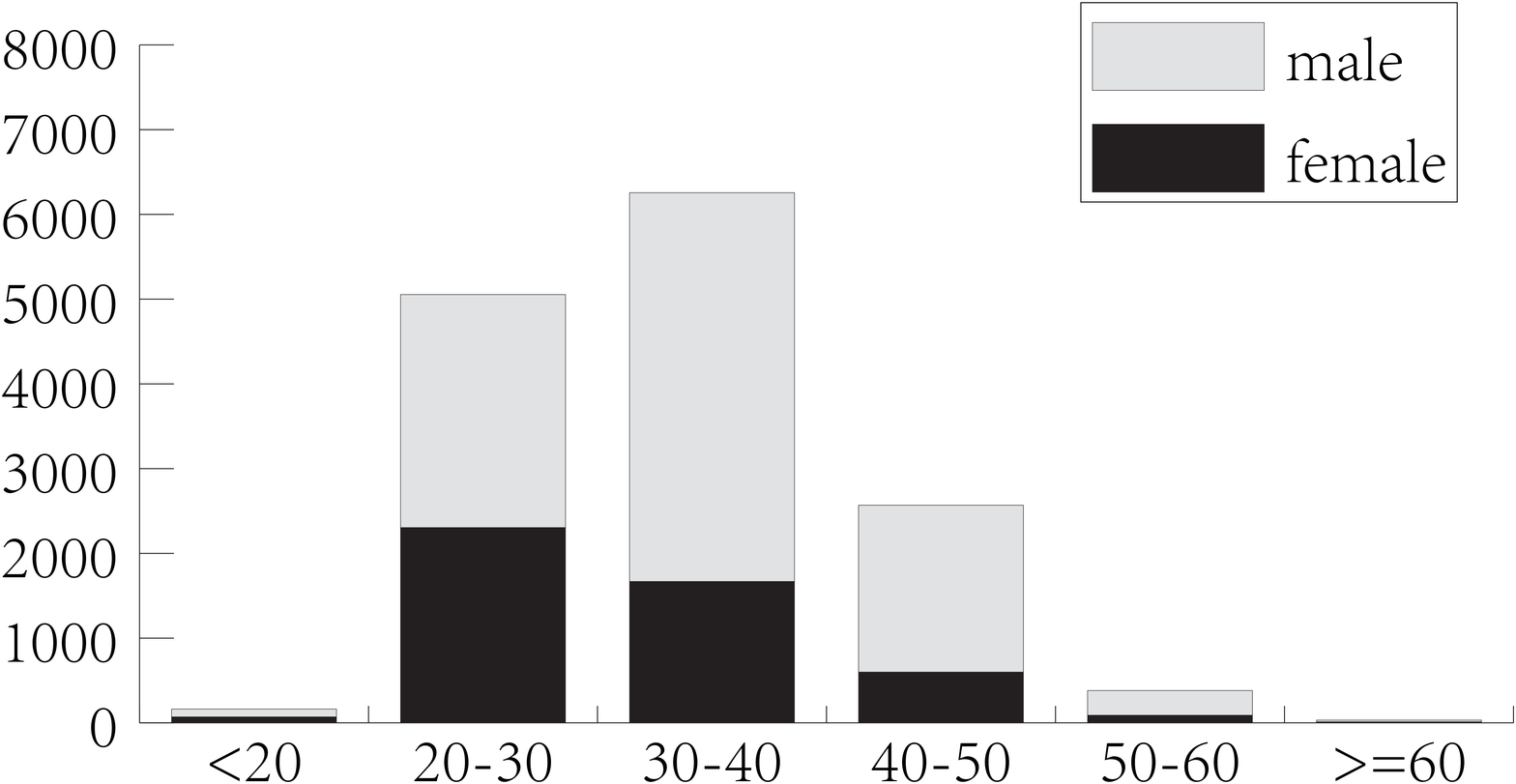}
}
\quad
\subfigure []{
\includegraphics[width=4cm]{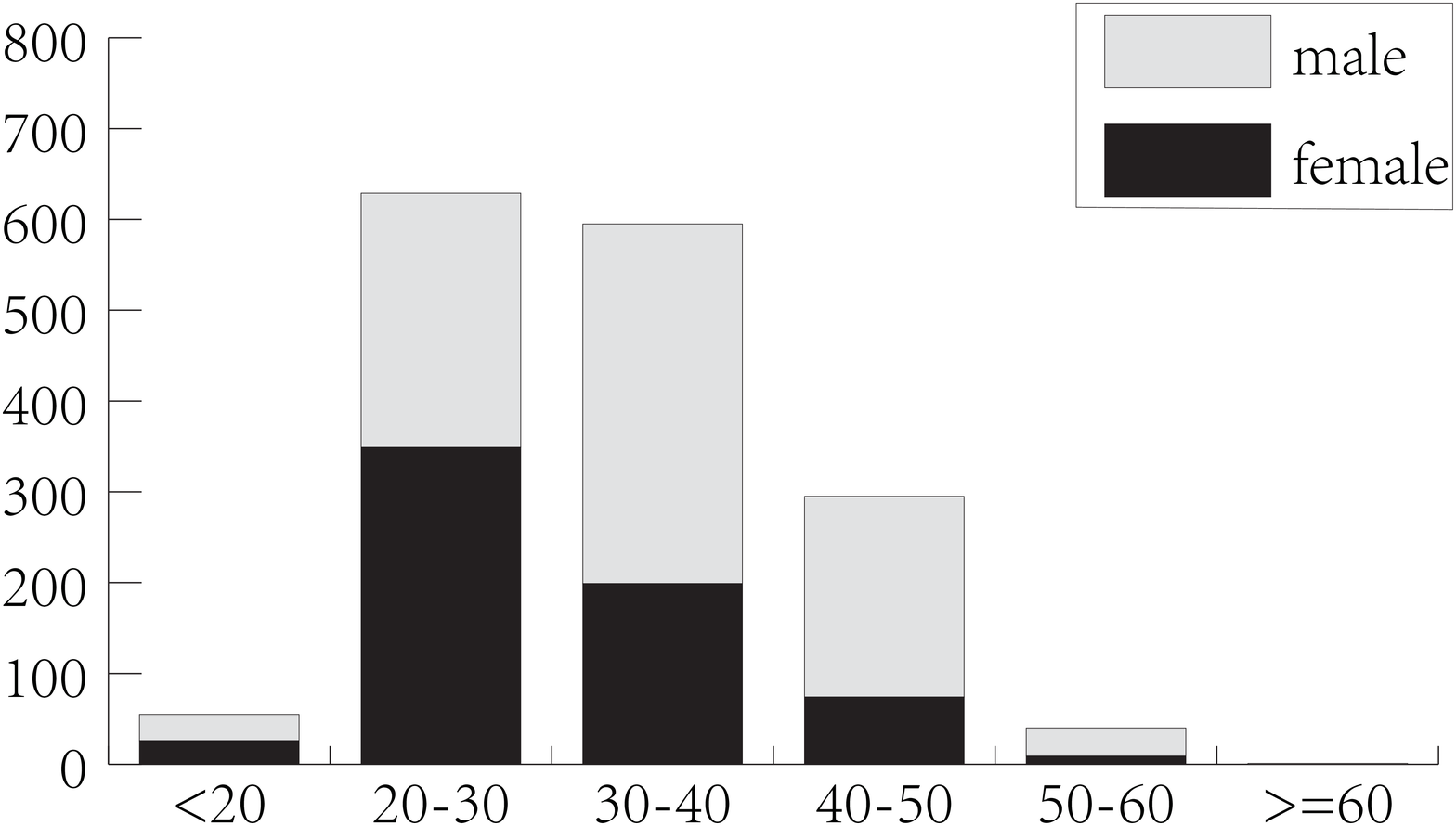}
}
\caption{The age distribution of HEU Emotion. (a) HEU-part1 (b) HEU-part2}
\label{fig5}
\end{figure}

The HEU Emotion database contains 19,004 video clips, which is divided into two parts: HEU-part1 and HEU-part2. HEU-part1 was divided into three parts, 80\% for training, 10\% for verification, and 10\% for testing. The concrete number of each category is presented in Tables \ref{table2} and \ref{table3}. Table \ref{table2} shows all the data of HEU-part1, while Table \ref{table3} only shows the data that face can be detected in HEU-part1. From the comparison of Tables \ref{table2} and \ref{table3}, it can be observed that faces are difficult to be detected or too small to judge their emotions in some clips. However, the emotional state of the characters can still be judged by their body postures (such as waving his arms when angry, holding his head with his hands when he is afraid, dancing when he is happy, etc.). This illustrates the necessity of body posture emotional data. The characters in the video clips include Asians, Africans, and Caucasians. We used DEX \cite{rothe2015dex} to estimate age and gender in facial data. According to the results, 67\% of faces are male, and women account for 33\%. Through age estimation, the average age of the male face is 34.67, and that of women is younger, 28.94 years old. In particular, the histogram of the number of faces in age ranges [0,20),[20,30),[30,40),[40,50),[50,60),[60,-]are depicted in Fig. \ref{fig5}a.

We used the same statistical method for HEU-part2. Drawing on the division of AFEW, HEU-part2 was divided into two parts, 65\% as the training set and 35\% as the validation set. Frames of HEU-part2 all contain facial expressions, the statistics of the categories are shown in Table \ref{table4}. Age and gender estimates were also performed on face data using DEX. Men account for 58\%, and women account for 42\%. The detailed age distribution is shown in Fig. \ref{fig5}b. Estimated by DEX, both datasets are male-majority, with ages concentrated at [20,50]. The result may be affected by datasets, which are used to train model weight. The actual gender and age distribution may be more balanced than that shown in Fig. \ref{fig5}. In addition, we performed face matching using the face recognition program in dlib \cite{kazemi2014one}. The number of subjects included in HEU Emotion is 9,951, of which HEU-part1 is 8,984 and HEU-part2 is 967.

\begin{figure}[htbp]
\centering
\includegraphics[width=8.5cm]{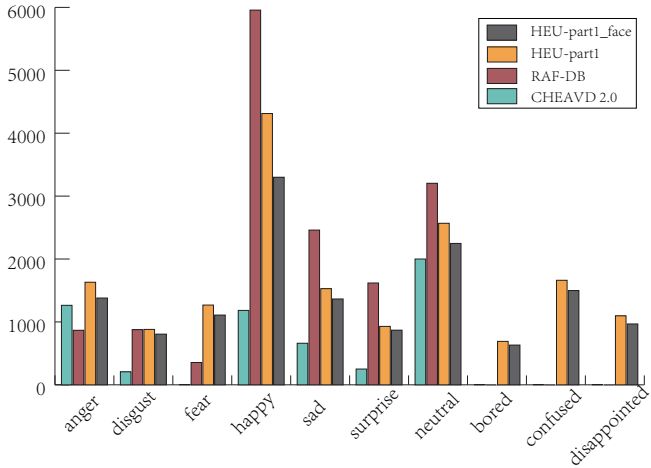}
\caption{The comparison between HEU-part1, RAF-DB and CHEAVD 2.0}
\label{fig6}
\end{figure}

\begin{figure}[htbp]
\centering
\includegraphics[width=8.5cm]{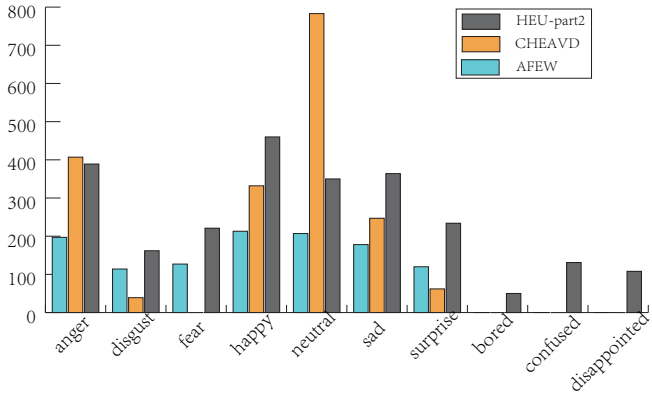}
\caption{The comparison between HEU-part2, AFEW and CHEAVD}
\label{fig7}
\end{figure}

We also compared HEU-part1 with CHEAVD2.0 and RAF-DB, as shown in Fig.\ref{fig6}. RAF is a facial expression dataset in the real environment which is widely used at present, and it has a relatively large amount of data. CHEAVD2.0 is a large multimodal emotion dataset. We compare these two datasets with HEU-part1. The amount of other emotions is similar or much higher than RAF-DB except for happy, anger and surprise. There are more clips of happy and neutral emotions in all three datasets. It shows that uneven distribution is a shared problem among most datasets. Category distribution of HEU-part1 is comparatively more balanced than others. The data sources of HEU-part2 are the same as those of AFEW. In Figure \ref{fig7}, HEU-part2 is compared with AFEW and CHEAVD. We only know the emotional categories of the CHEAVD training set and test set, so Figure \ref{fig7} shows the sum of the training set and test set of CHEAVD. It can see from Fig.\ref{fig7} that the number of all categories of HEU-part2 is higher than that of AFEW. In CHEAVD, neutral emotion data account for about half of the total sample, and the number of other hard-to-identify emotions (disgust,fear, and so forth.) is minimal. Although the number of some categories is smaller than CHEAVD, HEU-part2 data is relatively more balanced, especially when only seven basic emotions are evaluated.

\begin{figure}[htbp]
\centering
\subfigure []{
\includegraphics[width=2cm]{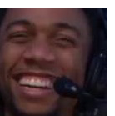}
}
\quad
\subfigure []{
\includegraphics[width=2cm]{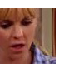}
}
\quad
\subfigure []{
\includegraphics[width=2cm]{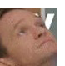}
}
\quad
\subfigure []{
\includegraphics[width=2cm]{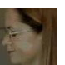}
}
\quad
\subfigure []{
\includegraphics[width=2cm]{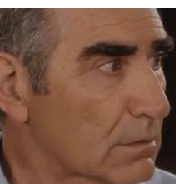}
}
\quad
\subfigure []{
\includegraphics[width=2cm]{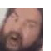}
}
\quad
\subfigure []{
\includegraphics[width=2cm]{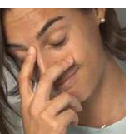}
}
\quad
\subfigure []{
\includegraphics[width=2cm]{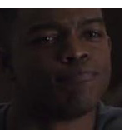}
}
\caption{The facial expressions of HEU Emotion in many different situations. (a) Look at the front horizontal (b) Overlook (c) Look up (d) Left face (e) Right face (f) Low resolution (g)Partial occlusion (h) Low light }\label{fig8}
\end{figure}

Facial expressions in the HEU Emotion dataset contain multiple perspectives (top, bottom, and horizontal), multiple angles (front, side), partial occlusion, and multiple resolutions. The intensity of expression in some videos will be transforming; for example, the expression starts with no emotion and then gradually increases the emotion to the peak. Figure \ref{fig8} displays some samples of the expression described above.

\begin{table*}[htbp] %开始一个表格environment，表格的位置是h,here。
\newcommand{\tabincell}[2]{\begin{tabular}{@{}#1@{}}#2\end{tabular}}
\centering
\caption{Number of annotated videos of HEU-part1 in each category} %显示表格的标题
\label{table2}
\begin{tabular}{ccccccccccc} %设置了每一列的宽度，强制转换。

\hline
%\toprule[0.8pt]
%Format & Extension & Description \\ % 用&来分隔单元格的内容 \\ 表示进入下一行
%\hline %画一个横线，下面的就都是一样了，这里一共有4行内容
%
& Anger& Bored& Confused& Disappointed& Disgust& Fear& Happy& Neutral& Sad& Surprise\\
\hline
Train& 1284& 551& 1330& 919& 690& 1012& 3330& 1985& 1202& 712\\
%\hline
Val& 170& 69& 166& 104& 96& 126& 492& 293& 164& 108\\
%\hline
Test& 177& 71& 166& 74& 95& 130& 491& 290& 163& 109\\
\hline

\end{tabular}
\end{table*}

\begin{table*}[htbp] %开始一个表格environment，表格的位置是h,here。
\newcommand{\tabincell}[2]{\begin{tabular}{@{}#1@{}}#2\end{tabular}}
\centering
\caption{Number of annotated videos of HEU-part1 (Face) in each category}\label{table3} % 显示表格的标题
\begin{tabular}{ccccccccccc} %设置了每一列的宽度，强制转换。

\hline
%\toprule[0.8pt]
%Format & Extension & Description \\ % 用&来分隔单元格的内容 \\ 表示进入下一行
%\hline %画一个横线，下面的就都是一样了，这里一共有4行内容
%
& Anger& Bored& Confused& Disappointed& Disgust& Fear& Happy& Neutral& Sad& Surprise\\
\hline
Train& 1094& 503& 1199& 807& 628& 880& 2534& 1740& 1066& 665\\
%\hline
Val& 150& 65& 147& 97& 89& 115& 423& 266& 148& 103\\
%\hline
Test& 136& 64& 153& 64& 89& 114& 343& 241& 151& 101\\
\hline
\end{tabular}
\end{table*}
\begin{table*}[htbp] %开始一个表格environment，表格的位置是h,here。
\newcommand{\tabincell}[2]{\begin{tabular}{@{}#1@{}}#2\end{tabular}}
\centering
\caption{Number of annotated videos of HEU-part2 in each category }\label{table4} %显示表格的标题

\begin{tabular}{ccccccccccc} %设置了每一列的宽度，强制转换。

\hline

& Anger& Bored& Confused& Disappointed& Disgust& Fear& Happy& Neutral& Sad& Surprise\\
\hline
Train& 252& 32& 85& 70& 101& 143& 299& 227& 236& 158\\
%\hline
Val& 137& 18& 46& 38& 61& 78& 161& 123& 128& 85\\

\hline
\end{tabular}
\end{table*}


\section{Baseline}\label{sec:baseline}
\begin{figure*}[htbp]
\centering
\includegraphics[width=15cm]{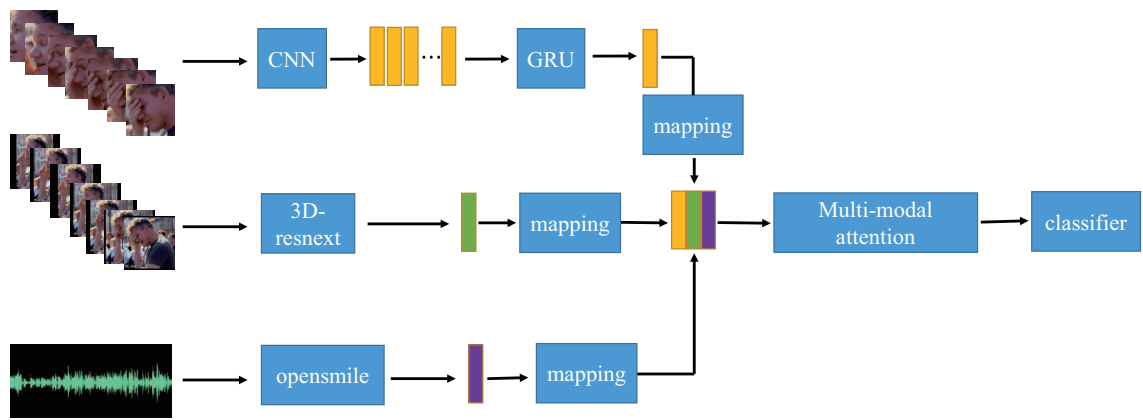}
\caption{The flowchart of our multi-modal emotion recognition method. It begins with extracting raw features of face, body, and audio. The raw features of face are transformed into video-level features by a GRU module. Then they are transformed to the same length by feature mapping. Different modal features are combined by a multi-modal attention module, and then fed to a classifier.}
\label{fig9}
\end{figure*}

In this section, four baselines are proposed for multi-modal emotion recognition for the HEU Emotion dataset. Video treatment flowsheet takes videos as input, extracts single-modal features and finally completes emotional recognition through a variety of fusion methods. The flowchart is shown in Fig. \ref{fig9}.

We designed four challenging benchmark experiments. (1) To expedite the transformation of research from the laboratory condition to the real environment, we conducted comparative experiments on the small popular CK+ dataset \cite{lucey2010extended}. (2) In order to prove the validity of our database, the comparative experiments were done with AFEW dataset, and the baseline of facial expression recognition was given. (3) The HEU Emotion was trained and tested by deep learning method. Moreover, baselines of body posture emotion and speech emotion were given. (4) Finally, the three modalities were merged in various ways, and the baseline of multi-modal emotion recognition was obtained.

\subsection{multi-modal attention mechanism}\label{subsec:MMA}
\begin{figure*}[htbp]
\centering
\includegraphics[width=15cm]{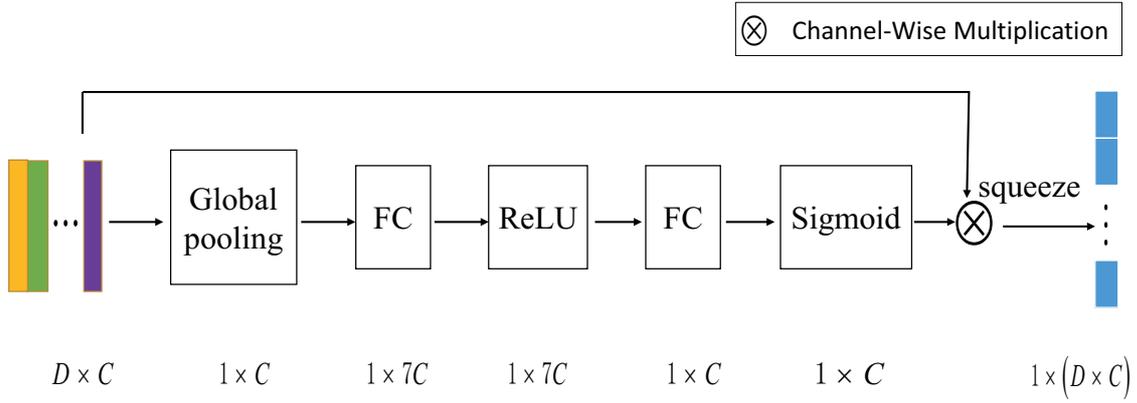}
\caption{ Multi-modal attention module}
\label{fig10}
\end{figure*}

Our dataset contains multiple modals of data. For one video, not every type of data plays a positive role in the final judgment. The judgment of human emotion is mainly based on facial expression, followed by the pronunciation and intonation of speech. The movement of the body can also assist the judgment to a certain extent. But it is worse than the other two kinds of information. All in all, the proportion of the data in the dataset to the judgment of the final emotion category is different. We propose a multimodal attention module (MMA) to adaptively adjust the proportion of different modal features according to their contribution to the classification results. The specific structure of the multimodal attention module is shown in Fig.\ref{fig10}.

First of all, we spliced the video-level features of each modality to get the input attention of the multimodal attention module $X \in {\mathbb{R}^{D \times C}}$, where $D$ represented the dimension of the video-level feature, and $C$ represented the number of input modes, which was called the number of channels. We used the global average pooling operation to compress along the dimensions of video-level features, turning each one-dimensional feature into a real number, which to some extent had a global receptive field and represented the global distribution of responses on each channel. Notice that the number of channels $C$ is constant.
\begin{equation}
{y_c} = \frac{1}{D}\sum\limits_{i = 1}^D {x_c^i}
\end{equation}
where ${y_c}$ represents the real number obtained by global average pooling on the $c$-th channel. $x_c^i$ represents the $i$ th element on the $c$-th channel. Then we use two full connection layers (FCs) to model the correlation between channels. In the first full connection layer, we increase the dimension of the channel to 7 times of the input. After being activated by the activation function ReLU, we reduce the channel to the original dimension through the second full connection layer. The use of two FC and ReLU makes the nonlinearity better and can better fit the complex correlation between channels. Generally speaking, the number of modes in multimodal fusion is not very large, and the channel is upgraded in order to improve the learning ability and information processing ability of the network. Furthermore, it will not bring too many parameters and calculations. Then a $Sigmoid$ function is used to normalize the resulting attention weight to between 0 and 1. Therefore, the output of the $Sigmoid$ function represents the importance of each channel after feature selection.
\begin{equation}
z = \sigma ({W_2}\delta ({W_1}y))
\end{equation}
where ${W_1} \in {\mathbb{R}^{c \times 7c}}$ represents the parameter of the first fully connected layer, where 7 is the scaling ratio. ${W_2} \in {\mathbb{R}^{7c \times c}}$ represents the parameters of the second fully connected layer. The correlation between modeling channels is displayed by learning the parameter $W$ of the network.

Then the original feature was re-calibrated in the channel dimension by weighting the previous feature channel by multiplication.
\begin{equation}
{u_c} = {z_c} \cdot {x_c}
\end{equation}
where ${u_c}$ represents the feature on the $c$-th channel after recalibration.

Finally, the feature map after feature selection was compressed into a vector and fed to the classifier.

\textbf{Discussion} MMA is inspired by the channel attention module SENet \cite{hu2018squeeze}. The main difference is that SENet captures the importance of each channel in the feature extraction process. The input here comes from the convolution results of different convolution kernels. The inputs of MMA are features from different modalities, and each feature is spliced as a channel. Then the problem of different effects of each modality on the final result in multi-modalities is transformed into modeling the correlation between each channel. Through the automatic learning mode obtain the importance of each feature channel. Then according to this importance the useful features are enhanced and the features that are not useful are suppressed for the current task. Another difference is that SENet is averagely pooled on a two-dimensional feature graph, while our input features are multiple one-dimensional vectors, so our global average pooling is carried out in one-dimensional vectors.

\subsection{Comparison with CK+ Dataset}\label{subsec:comparsion with CK+ dataset}
In the experiments from the laboratory environment to the real world, the two parts of HEU-Emotion were made. We conducted comparative experiments using the most popular dataset CK+. Firstly, the facial information of CK+ dataset was extracted. Then the dataset was split into training set and test set in a ratio of 4:1 for each type of expression. Furthermore, the last five pictures of each folder were chosen as samples. The validation set and test set of HEU-part1 were combined together to be used as the test set, so the proportion of training and test sets was 4:1. The ratio of HEU-part2 remained unchanged at 0.65:0.35. In the same way, five pictures of each folder were supposed to form an experimental sample. Taking into account of the small dissimilarity between adjacent frames, the following sampling method was adopted: if the total number was less than 5, all were selected; if less than 10, the first 5 were taken; if less than 14, one picture was taken every interval; If there were less than 18, one picture was taken at intervals of two frames; the remaining interval was taken as 3 frames.

In the baseline experiment, hand-crafted features extracted by conventional machine learning methods were used. For Local Binary Pattern (LBP)\cite{ahonen2004face}, the images were resized to \begin{math} 128 \times 128 \end{math} pixels as input and then divided into \begin{math} 8 \times 8 \end{math} pixel blocks. The LBP descriptor applied the Uniform Pattern and generated 3,776-dimensional feature vectors for each image. The Histogram of Oriented Gradient (HOG) features \cite{dalal2005histograms} also used images of \begin{math} 128 \times 128 \end{math} pixels as input. The HOG features utilized shape-based segmentation dividing the image into \begin{math} 16 \times 16 \end{math} pixel blocks of four \begin{math} 8 \times 8 \end{math} pixel cells with no overlapping. We got 8,100-dimensional HOG feature vectors for each image by setting nine bins. For Gabor wavelet \cite{liu2002gabor}, we adjusted face images to the size of \begin{math} 16 \times 16 \end{math} pixels and implemented 128 Gabor filters in 16 spatial scales and eight directions. Each image was divided into \begin{math} 10 \times 10 \end{math} blocks, each block was filtered with 128 filters, and finally, the feature vectors of 12,800 dimensions were obtained. After feature extraction using LBP, HOG and Gabor wavelets, Support Vector Machines (SVM)\cite{suykens2001support} was used as classifiers for classification.

\begin{table}[h] %开始一个表格environment，表格的位置是h,here。
\newcommand{\tabincell}[2]{\begin{tabular}{@{}#1@{}}#2\end{tabular}}
\centering
\caption{Comparison with CK+ using LBP, HOG and Gabor wavelet features}\label{table5} % 显示表格的标题

\begin{tabular}{cccc}
		\hline
		\multirow{2}{*}{Method}    & \multicolumn{3}{c}{Accuracy(\%)}              \\ \cline{2-4}
		& CK+        & HEU-part1        & HEU-part2 \\ \hline
		\multicolumn{1}{c}{LBP} & 76.22& 28.06& 35.88                         \\ %\hline
		\multicolumn{1}{c}{HOG} & 83.54& 27.34& 29.63                          \\ %\hline
		\multicolumn{1}{c}{Gabor} & 78.28& 23.82& 24.54                          \\ \hline

	\end{tabular}
\end{table}

 It can be seen from the second column of the data in Table \ref{table5} that the extracted LBP, HOG and Gabor features are valid on the face data. When they were used on the HEU Emotion dataset, the accuracy rate is only over 20\%, with a maximum of 35.88\%. The main reason why HEU Emotion performs poorly is that the actual conditions are more challenging than the laboratory environment. The illumination change and partial occlusion problems increase the difficulty of recognition. What is more, the result also indicates that the effectiveness of transferring the method developed in the laboratory environment to the real condition is not guaranteed. In order to further develop the actual application of affective computing, it is necessary to establish a large-scale emotional dataset in the real environment. As the most massive dynamic, temporal multimodal emotion recognition database, HEU Emotion will undoubtedly promote the advancement of emotion recognition.

\subsection{Cross-dataset Experiment}\label{subsec:cross-dataset experiment}
Deep learning has made breakthroughs in many application areas of artificial intelligence such as machine vision, speech recognition, and natural language processing. In recent years, the methods proposed in the emotion recognition task are also based on deep learning. From the best papers of the EmotiW Challenge \cite{fan2016video,hu2017learning,liu2018multi}, it can be seen that the deep learning methods are very advantageous to solve emotion recognition problems in the real environment. To demonstrate the validity of the HEU Emotion dataset, we compared it with the AFEW dataset. For deep learning methods, we took classic convolutional neural network models such as VGG \cite{simonyan2014very}, Resnet \cite{he2016deep}, Densenet \cite{huang2017densely}, and SE-inception \cite{hu2018squeeze} as feature extraction networks, where VGG was the best model in the paper of the EmotiW 2016 Challenge winner \cite{fan2016video}. The HEU Emotion is an emotion recognition dataset of sequencial task. After performing feature extraction of a single frame, we used the GRU to process the features further and finally sent them to the classifier to get the emotional category of the clip. Before training, face image was resized to \begin{math} 229 \times 229 \end{math} pixels. In the training phase, image data enhancement operations such as random cropping and horizontal flipping were performed, and the input image was randomly cropped to \begin{math} 224 \times 224 \end{math} pixels. Sixteen pictures were randomly selected from a video sequence as inputs in the training process. If the entire video sequence was less than 16, the last picture was copied until 16 pictures were satisfied. In the validation and test phase, 16 consecutive frames in the video were taken. If the total number of frames was more than 16, it was taken as the second group, and so on until the entire pictures were taken, and 8 frames in the adjacent group were overlapped. When multiple sets of input were taken, the average of their outputs was fed into softmax to give the final result. As a result, a video was only available in one emotional tag.

Convergence was particularly slow when using HEU Emotion to train the model directly. Considering that using pre-training models can speed up the convergence \cite{levi2015emotion,ng2015deep,peng2016towards}, we trained the model of the convolutional neural network in the popular non-laboratory static expression database FER2013 \cite{goodfellow2013challenges} to obtain the parameters. When training the AFEW and HEU Emotion datasets, all parameters except that of the classifier were loaded and then globally fine-tuned. The network was trained with 500 epochs and used the batch size of 16. Since a pre-training model was used, the initial learning rate was set to 0.0001. The optimizer used a stochastic gradient descent (SGD) with a fixed learning rate.

\begin{table*}[htbp] %开始一个表格environment，表格的位置是h,here。
\newcommand{\tabincell}[2]{\begin{tabular}{@{}#1@{}}#2\end{tabular}}
\centering
\caption{The baseline of HEU Emotion face data} % 显示表格的标题
\label{table6}

\begin{tabular}{ccccc}
		\hline
		\multirow{2}{*}{Method}    & \multicolumn{4}{c}{Accuracy(\%)}              \\ \cline{2-5}
		& AFEW(val) & HEU-part1(val)        & HEU-part1(test)        & HEU-part2(val) \\ \hline
VGG+GRU \cite{simonyan2014very} & 50.28& 47.53& 41.19& 51.03\\
%\hline
Resnet+GRU \cite{he2016deep}& 50.13& 44.10& 40.35& 43.31\\
%\hline
Densenet+GRU \cite{huang2017densely}& 38.07& 40.05& 39.21& 42.41\\
%\hline
SE-inception+GRU \cite{hu2018squeeze}& 35.23& 42.25& 37.68& 43.34\\
\hline
\end{tabular}

\end{table*}

The partition of the HEU Emotion dataset is described in Sect. \ref{sec:statistics}. According to the distribution of EmotiW2019 dataset, AFEW dataset is divided into training set (773) and verification set (383). The second column of Table \ref{table6} is the result of the verification set trained on the AFEW dataset. It can be seen that the classical network is effective for facial expression feature extraction. In particular, the features extracted by VGG are fed into GRU to achieve the highest accuracy of 50.28\%. The same algorithm has different results on our HEU Emotion dataset. For HEU-part1, it reached 47.53\% on the verification set and 41.19\% on the test set. The accuracy of HEU-part2 is 51.03\%. Combining other methods, compared with AFEW data, we can see that the HEU Emotion dataset is practical and the data sample complexity of HEU emotion is higher. HEU Emotion is a more challenging database, and the emotion recognition algorithm proposed based on it can be closer to the real application.

\begin{table}[h]
%开始一个表格environment，表格的位置是h,here。
\newcommand{\tabincell}[2]{\begin{tabular}{@{}#1@{}}#2\end{tabular}}
\centering
\caption{The Experiments of cross-datasets} % 显示表格的标题
\label{table7}
\begin{tabular}{cccc} %设置了每一列的宽度，强制转换。

\hline
%\toprule[0.8pt]
%Format & Extension & Description \\ % 用&来分隔单元格的内容 \\ 表示进入下一行
%\hline %画一个横线，下面的就都是一样了，这里一共有4行内容
%
Train set& \tabincell{c}{Number of\\ categories} & Test set& Accuracy(\%)\\
\hline
\multirow{5}*{HEU-part2} & 10& AFEW(train)& 16.50\\
%\cline{2-4}
 & 10& AFEW(val)& 10.36\\
 %\cline{2-4}
 & 7& HEU-part2(val)& 53.17\\
 %\cline{2-4}
 & 7& AFEW(train)& 48.80\\
 %\cline{2-4}
 & 7& AFEW(val)& 41.19\\
\hline
\multirow{6}*{HEU-part1} & 10& AFEW(train)& 16.19\\
%\cline{2-4}
 & 10& AFEW(val)& 13.63\\
%\cline{2-4}
& 7& HEU-part1(val)& 58.40\\
%\cline{2-4}
& 7& HEU-part1(test)& 51.46\\
%\cline{2-4}
& 7& AFEW(train)& 56.26\\
%\cline{2-4}
& 7& AFEW(val)& 42.33\\
\hline
\end{tabular}

\end{table}

Table \ref{table7} shows the experimental results of cross-dataset testing. We use VGG+GRU, the most effective method shown in Table \ref{table6}, to experiment across datasets. The first two columns in Table \ref{table7} show the datasets used for training and the corresponding number of categories, the third column shows the test sets used, and the last column shows the accuracy on the corresponding test sets. Firstly, the parameter model trained on the two-part dataset of HEU Emotion is tested on the training and verification dataset of AFEW. Emotion categorie of HEU Emotion is 10.  Consequently, the model parameter can not be directly applied to AFEW. We replaced 10 classifiers with classifiers trained by AFEW. From the first and second rows of Table \ref{table7}, we can see that the parameters trained with all 10 types of data from HEU-part2 have poor results on the AFEW dataset. After that, we used the basic seven types of emotion data in HEU-part2 to train and got the accuracy of 53.17\%. When the parameters obtained from seven kinds of training data are tested on the training set of AFEW, the accuracy is 48.80\%, and the accuracy on the verification set is 41.19\%. The reason why the accuracy of the training set is higher than that of the verification set should be that the number of samples on the training set is twice that of the verification set. The larger the amount of data, the smaller the chance, the results can reflect more exactly the real performance of the algorithm. Compared with the accuracy of AFEW in Table \ref{table6}, that of HEU-part2 (7 class) is slightly lower. But it reveals that the sample distribution of HEU-part2 and AFEW is very similar to the annotation standard. We used the same authentication method for HEU-part1. From the six or seven rows in Table \ref{table7}, we can see that the feature distribution obtained by 10 classification is different from that of 7 classification, and the result is not ideal. In order to further explain that the number of categories of HEU-part1 in the training process will affect the final feature distribution. Finally, we chose the seven basic categories of the HEU-part1 training set for training. The results containing only seven categories are 58.40\% in verification set and 51.46\% in test set respectively. Then the parameters were applied to the AFEW. The accuracy of AFEW training set is 56.26\%, and that of verification set is 42.33\%.

As can be seen from the analysis of Tables \ref{table6} and \ref{table7}, when there are only 7 types of expressions in the HEU Emotion, the results on the validation and test sets are better. The addition of the other three types of expressions reduces the overall accuracy. To verify the influence of the other three emotions, the diagonal values of the normalized confusion matrix on the HEU Emotion are listed in Table \ref{table8}.

It can be observed from Table \ref{table8} that except for bored, confused, and disappointed emotions, the accuracy of most other categories is slightly higher than that of 10, especially anger and neutral. The results indicates that the addition of categories does affect the distribution of final features, especially bored, which is highly similar to neutral data.

\begin{table*}[tbp] %开始一个表格environment，表格的位置是h,here。
%\begin{threeparttable}
\newcommand{\tabincell}[2]{\begin{tabular}{@{}#1@{}}#2\end{tabular}}
\centering
\caption{Facial expression recognition performance of VGG on HEU Emotion. The metric is the diagonal value of the normalized confusion matrix} % 显示表格的标题
\label{table8}
\begin{tabular}{ccccccccccc} %设置了每一列的宽度，强制转换。

\hline
Dataset& Anger& Bored & Confused& \tabincell{c}{Disappoi \\ -nted}& Disgust& Fear& Happy& Neutral& Sad& Surprise\\
\hline
HEU-part1(val) & 0.39& 0.12& 0.34& 0.10& 0.24& 0.27& 0.81& 0.56& 0.50& 0.38\\
%\hline
HEU-part1(test)& 0.27& 0.17& 0.32& 0.17& 0.13& 0.25& 0.78& 0.48& 0.43& 0.25\\
%\hline
HEU-part1(val,7)& 0.44& -& -& -& 0.34& 0.30& 0.83& 0.66& 0.51&0.32\\
%\hline
HEU-part1(test,7)& 0.38& -& -& -& 0.18& 0.31& 0.78& 0.63& 0.49& 0.22\\
%\hline
HEU-part2(val)& 0.65& 0.22& 0.43& 0.42& 0.25& 0.45& 0.73& 0.46& 0.39& 0.42\\
%\hline
HEU-part2(val,7)& 0.72& -& -& -& 0.28& 0.40& 0.74& 0.40& 0.48& 0.42\\
\hline
\end{tabular}
%\begin{tablenotes}
%\footnotesize
%     \item[*] Facial Expression Recognition Performance of VGG on HEU Emotion
%   \end{tablenotes}
%  \end{threeparttable}
\end{table*}
\subsection{The Baselines of other Modalities}\label{subsec:the baselines of other Modalities}
In addition to the baseline for facial expression recognition, we also used the method of deep convolutional neural network (CNN) to give baselines for another two modalities. Body postures emotion recognition and human behavior recognition were both judged by human body movements, but the target domains were different. Therefore, we applied the methods of human behavior recognition research \cite{hara2018can,simonyan2014two,zhu2018hidden} in recent years to posture emotion recognition. To accelerate the convergence speed, we used the weight of these models on action recognition as the pre-training model. Many networks were given in \cite{hara2018can}. We chose the best-performing model ResNeXt-101. Each image was resized to \begin{math} 112 \times 112 \end{math} pixels before entering the network. In the training phase, we took 16 consecutive frames as one input, and the index of the starting image was random. When the index of the ending image exceeded the range, the last image was reproduced to 16 frames. In the test phase, 16 consecutive frames were input, besides here all the pictures were taken in sequence. The batch size was 3, the initial learning rate of HEU-part1 was 0.01, and HEU-part2 was 0.1. The learning rate was reduced to one-tenth of the previous one every 30 times. In addition, we chose ResNet101 as the spatial model in \cite{simonyan2014two}. First, images were resized into \begin{math} 224 \times 224 \end{math} pixels. During the training process, three frames were randomly selected from each video, and all frames were used in the testing stage. The initial learning rate was set at 0.0005, and the learning rate decreased by 0.1 times for every 50 batches. The optimizer used the Stochastic Gradient Descent method (SGD) with momentum. For \cite{zhu2018hidden}, we also used the spatial model Resenet 152 to train with a random frame image in each video, and verified with a single-frame image in the middle of the video. The batch size was set to 25. The initial learning rate was 0.01, and the learning rate per 100 batches was attenuated by 0.1 times.

\begin{table}[h] %开始一个表格environment，表格的位置是h,here。
\newcommand{\tabincell}[2]{\begin{tabular}{@{}#1@{}}#2\end{tabular}}
\centering
\caption{The Baseline of emotion recognition based on body posture} % 显示表格的标题
\label{table9}
\begin{tabular}{cccc}
		\hline
		\multirow{2}{*}{Model}    & \multicolumn{3}{c}{Accuracy(\%)}              \\ \cline{2-4}
		& HEU-part1(val)        & HEU-part1(test)        & HEU-part2(val) \\ \hline
		\multicolumn{1}{c}{ResNeXt101 \cite{8578783}} & 35.02& 33.02& 24.46                         \\ %\hline
		\multicolumn{1}{c}{ResNet101 \cite{simonyan2014two}} & 43.17& 42.86& 39.01                          \\ %\hline
		\multicolumn{1}{c}{ResNet152 \cite{zhu2018hidden}} & 34.84& 31.66& 28.48                          \\ \hline

	\end{tabular}
\end{table}

Table \ref{table9} shows the baseline of body posture emotion recognition of HEU Emotion database. ResNeXt-101 uses 3D convolution kernels to process sequencial data, and another two methods are based on single-frame images. Compared with the baseline of the facial expressions given in Table \ref{table6}, when the accuracy of facial expression recognition in the HEU-part1 test dataset achieves 41.19\%, the accuracy of the posture emotion recognition can reach 33.02\%. It proves that posture emotion recognition is also a meaningful way to judge emotions and also shows the effectiveness of posture emotion data of HEU Emotion dataset. Accuracy of HEU-part1 is higher than that of HEU-part2, but the opposite is exact in facial expression recognition. Considering that HEU-part1 has more videos of posture emotions, in other words, it contains the whole body or movements of a broader range. It can also be seen from the quantitative comparison in Table \ref{table2} and Table \ref{table3}. HEU-part2 is mostly close-shot, which contains slightly worse body information. Although HEU-part2 includes less posture information, it has more emotional speech information.

In order to get the baseline of emotional speech, we first calculated 16 low level descriptors (LLDs), which consisted of zero-crossing rate, energy square root, pitch frequency (F0), signal-to-noise ratio (HNR), MFCC1-12. Then, we got 32 LLDs by calculating the first-order difference of these 16 LLDs. These audio features can be extracted by openSMILE toolkit \cite{eyben2013recent} based on INTERSPEECH 2009 audio template \cite{schuller2009interspeech}. Each video clip was computed to obtain 384-dimensional features, and then classified by SVM.

We also gave another baseline using raw speech information and a deep convolutional neural network. The sampling rate of speech was 16K, that was, the speech in one second was a one-dimensional vector with a length of 16,000. The length of each speech sample was inconsistent.When the length was more than 20,000, the vector was randomly truncated to the length of 20,000, and when the length was less than 20,000, it was supplemented with 0. The vector of length 20,000 was sent to the one-dimensional convolution network (ResNet10) for training. During the test, each sample was cut into 20,000-dimensional vectors by a step size of 160, and multiple one-dimensional vectors can be gained. The test results were put to the vote to obtain the final prediction result of each sample.

\begin{table}[h] %开始一个表格environment，表格的位置是h,here。
\newcommand{\tabincell}[2]{\begin{tabular}{@{}#1@{}}#2\end{tabular}}
\centering
\caption{The baseline of speech emotion recognition} % 显示表格的标题
\label{table10}
\begin{tabular}{cc} %设置了每一列的宽度，强制转换。

\hline
%\toprule[0.8pt]
%Format & Extension & Description \\ % 用&来分隔单元格的内容 \\ 表示进入下一行
%\hline %画一个横线，下面的就都是一样了，这里一共有4行内容
%
Model& Accuracy(\%)\\
\hline
Opensmile+SVM \cite{eyben2013recent}& 37.15\\
%\hline
ResNet10 \cite{he2016deep}& 28.64\\
\hline
\end{tabular}
\end{table}

The accuracy of the speech emotion recognition on the HEU-part2 validation set is shown in Table \ref{table10}. It reveals that the audio features extracted by openSMILE can get a better classification result of 37.15\%. Compared with another two modalities, the speech emotion also achieves competitive results. Hence, the emotional speech data of HEU-part2 is also effective.

\subsection{Multi-modal Emotion Recognition}\label{subsec:multi-modal emotion recognition}
From the baselines of the three single modalities given in Sect. \ref{subsec:cross-dataset experiment} and Sect.\ref{subsec:the baselines of other Modalities}, single-modal emotion recognition can achieve excellent results. Compared with single modality, multi-modality can get higher accuracy by complementing each other. Our database contains faces, body postures, and speeches, which all play roles in the judgment of emotions. When a face is not available, the body posture can help to recognize emotion. In some cases, some emotional expressions do not have corresponding postures. Therefore, we need to perform a comprehensive analysis of the three modalities. We used the facial expression recognition model \cite{simonyan2014very}, the posture emotion recognition model \cite{8578783} and the speech emotion recognition model \cite{eyben2013recent} to perform fusion in various ways. We used three fusion methods: MMA, fusion proposed by \cite{ben2019block} and  late fusion \cite{yan2018multi}. During the forecasting phase of late fusion, each model predicted the probability that the sample belonged to the relevant emotion. Assign appropriate weights to each model to get the following probability representation:

\begin{equation}
p = w_1 \times p_{face0}+w_2 \times p_{audio}+w_3 \times p_{body}+w_4 \times p_{face1}
\end{equation}
Where $\sum_{i=1}^4w_i=1$, $ w_i $ is the weight of each model. The subscript of $p$ represents the corresponding modality, and the weight is $w_i=0$ when the corresponding modality is not available. We used a grid search strategy to get the weight.

\begin{table}[h]
	\centering
	\caption{Comparison to the state-of-art}
\label{table11}
	\begin{tabular}{cccc}
		\hline
		\multirow{2}{*}{Method}    & \multicolumn{3}{c}{Accuracy(\%)}              \\ \cline{2-4}
		& HEU-part1(val)        & HEU-part1(test)        & HEU-part2(val) \\ \hline
		\multicolumn{1}{c}{Block \cite{ben2019block}} & 47.79          & 41.52             & 51.34                          \\ %\hline
		\multicolumn{1}{c}{Late fusion \cite{yan2018multi}} & 48.08          & 42.03             & 53.26                          \\ %\hine
		\multicolumn{1}{c}{MMA} & 49.22          & 43.38             & 55.04                          \\ \hline

	\end{tabular}
\end{table}

\begin{table}[h]
	\centering
	\caption{The Results of different modalities using MMA.}
\label{table12}
	\begin{tabular}{cccc}
		\hline
		\multirow{2}{*}{Modal}    & \multicolumn{3}{c}{Accuracy(\%)}              \\ \cline{2-4}
		& HEU-part1(val)        & HEU-part1(test)        & HEU-part2(val) \\ \hline
		\multicolumn{1}{c}{Face} & 47.53          & 41.19             & 51.03                          \\ %\hline
		\multicolumn{1}{c}{Face+Audio} & -         & -             & 54.68                          \\ %\hline
		\multicolumn{1}{c}{All} & 49.22          & 43.38             & 55.04                          \\ \hline

	\end{tabular}
\end{table}
From the results in Tables \ref{table11} and \ref{table12}, we get the following observations. Our proposed multimodal fusion module MMA achieves better results than other methods. That is mainly due to the fact that MMA can adaptively adjust each mode on the classification results. Compared with the late-fusion fixed model weight, the way is more flexible and accurate. The HEU-part1 verification set increased by 1.69\% from 47.53\% to 49.22\%, and the test set increased from 41.19\% to 43.38\%, an increase of 2.19\%. The verification set of HEU-part2 has an accuracy of 51.03\% on the facial expression dataset, and the overall accuracy is improved by 4.01\% after being fused with speech and posture. It proves that multimodal information is essential for video-based emotion recognition. From Tables \ref{table6}, \ref{table9} and \ref{table10}, we can see that facial expressions have more advantages over other modes. The multi-modalities can make up for the impact of occlusion or other noise on facial expression recognition and achieve better recognition accuracy.

\section{Conclusion}\label{sec:conclusion}
Human emotion recognition is a challenging but meaningful research work. However, the existing multimodal emotional databases in the wild are small in scale, with a limited number of subjects, expressed in a single language, or few samples containing certain emotions. In this article, we collected and annotated a new multi-language multimodal video emotion database HEU Emotion (the number of subjects is 9,951). It consists of two parts: the bimodal database HEU-part1 (including facial expressions and body posture emotions) and multimodal database HEU-part2 (including facial expressions, body postures, and speeches). HEU Emotion was recorded in a multimodal synchronous way and can be directly used for multi-modal emotion recognition experiments. The videos of HEU Emotion were taken under uncontrollable natural conditions and included a variety of the change of multi-view face and body posture, local occlusion, illumination change, and changes in expression intensity. Our experiments proved that the algorithm trained on the lab-controlled databases were no longer suitable for emotion recognition tasks in the wild. Cross-dataset experiments with AFEW demonstrated that our database was superior to the AFEW. Finally, we performed multi-modal emotion recognition experiments using facial expressions, body postures, and emotional speeches. The evaluation indicators showed that it was beneficial to use multi-modality to process real-world video. To date, HEU Emotion has been the largest multimodal emotional recognition database in the natural environment. We hope that HEU Emotion can promote the development of multimodal emotion recognition and improve the performance of automatic affective computing systems in real-world applications.
%\begin{acknowledgements}
%\end{acknowledgements}


\bibliographystyle{spmpsci}      % mathematics and physical sciences
\bibliography{ref}   % name your BibTeX data base

%\begin{thebibliography}{}
% etc
%\end{thebibliography}

\end{document}